\documentclass[]{spie}  

\usepackage{amsmath,amsfonts,amssymb}
\usepackage{graphicx}
\usepackage{tabularx}
\usepackage{booktabs}
\usepackage{multirow}
\usepackage[colorlinks=true, allcolors=blue]{hyperref}

\title{GAN based Unsupervised Segmentation: \\Should We Match the Exact Number of Objects}

\author[a]{Quan Liu}
\author[b]{Isabella M. Gaeta}
\author[b]{Bryan A. Millis}
\author[b]{Matthew J. Tyska}
\author[a]{Yuankai Huo}
\affil[a]{Department of Electrical Engineering and Computer Science, Vanderbilt University, Nashville, TN, 37235, USA}
\affil[b]{Department of Cell and
Developmental Biology, Vanderbilt University, Nashville, TN, 37235, USA}

\authorinfo{Further author information: Quan Liu : E-mail: quan.liu@vanderbilt.edu\\
Contact Author: Yuankai Huo: E-mail: yuankai.huo@vanderbilt.edu}

\begin{document} 
\maketitle

\begin{abstract}
The unsupervised segmentation is an increasingly popular topic in biomedical image analysis. The basic idea is to approach the supervised segmentation task as an unsupervised synthesis problem, where the intensity images can be transferred to the annotation domain using cycle-consistent adversarial learning. The previous studies have shown that the macro-level (global distribution level) matching on the number of the objects (e.g., cells, tissues, protrusions etc.) between two domains resulted in better segmentation performance. However, no prior studies have exploited whether the unsupervised segmentation performance would be further improved when matching the exact number of objects at micro-level (mini-batch level). In this paper, we propose a deep learning based unsupervised segmentation method for segmenting highly overlapped and dynamic sub-cellular microvilli. With this challenging task, both micro-level and macro-level matching strategies were evaluated. To match the number of objects at the micro-level, the novel fluorescence-based micro-level matching approach was presented. From the experimental results, the micro-level matching did not improve the segmentation performance, compared with the simpler macro-level matching.
\end{abstract}

\keywords{Unsupervised learning, Segmentation, Synthesis, GAN}
\section{INTRODUCTION}
\label{sec:intro}  

Semantic segmentation is one of the central tasks in microscope image analysis, which segments targeting objects from background context~\cite{wu1995live}. Traditionally, the semantic segmentation was performed by unsupervised intensity based methods, such as watershed~\cite{pinidiyaarachchi2005seeded}, Gaussian mixture model (GMM)~\cite{ragothaman2016unsupervised}, graph-cut~\cite{lesko2010live} etc. In the past few years, the deep learning based methods have been increasingly popular in microscopy imaging, due to the superior accuracy and better generalizability~\cite{moen2019deep}. However, one of the major limitations in deep learning based semantic segmentation is the need of large-scale annotated images, which is not only tedious, but also resource intensive~\cite{zhang2017deeppap}. CycleGAN~\cite{zhu2017unpaired}, a breakthrough generative adversarial network (GAN)~\cite{goodfellow2014generative} was proposed recently, which shed light on semantic segmentation with minimal or even no manual annotation~\cite{huo2018adversarial,huo2018synseg,zhang2018translating,chen2019synergistic}.

Within the CycleGAN framework~\cite{zhu2017unpaired}, many previous studies have tackled the unsupervised semantic segmentation in microscopy imaging. Ihle et al.~\cite{ihle2019udct} proposed to use the CycleGAN framework to segment bright-field images of cell cultures, a live-dead assay of C.Elegans, and X-ray-computed tomography of metallic nanowire meshes. A similar approach was proposed by~\cite{gadermayr2019generative} for facilitating stain-independent supervised and unsupervised segmentation on kidney histology. DeepSynth~\cite{dunn2019deepsynth} was proposed to further extend the CycleGAN framework from 2D to 3D nuclear segmentation. Even though the CycleGAN based unsupervised segmentation approaches have shown decent performance on microscope images, very few studies have investigated the challenging sub-cellular microvilli segmentation with fluorescence microscopy imaging. The sub-cellular microvilli segmentation is challenging due to the highly overlapping and dynamic nature of such small sub-celluar objects~\cite{julio2008image,meenderink2019actin}. 

Different from Pix2Pix GAN~\cite{isola2017image}, which requires pixel-level matching between images across two domains, CycleGAN is able to perform image synthesis without paired images. However, the previous studies emphasized that the macro-level (global distribution level) matching on the number of objects between intensity images and simulated masks improved the segmentation performance~\cite{ihle2019udct}. That fact inspired us with the question that if the segmentation performance could be further improved by doing more careful matching than the macro-level. To answer the question, we propose a new micro-level matching (mini-batch level) strategy to match the rough number of objects across two domains when training the CycleGAN framework.

In this paper, we develop a deep learning based unsupervised semantic segmentation method for sub-cellular microvilli segmentation using fluorescence microscopy. Meanwhile, we evaluate the performance of micro-level matching strategy, which is enabled by the multi-channel nature of fluorescence images. The contributions of this study are three-fold: (1) We propose the first deep learning based unsupervised sub-cellular microvilli segmentation method; (2) We propose the micro-level matching to ensure the roughly same number of objects across two modalities within each mini-batch, without introducing extra human annotation efforts; (3) Comprehensive analyses are provided to evaluate the outcomes of different augmentation strategies when generating the simulated masks for unsupervised microvilli segmentation. 

\begin{figure}[t]
\begin{center}
\includegraphics[width=1\linewidth]{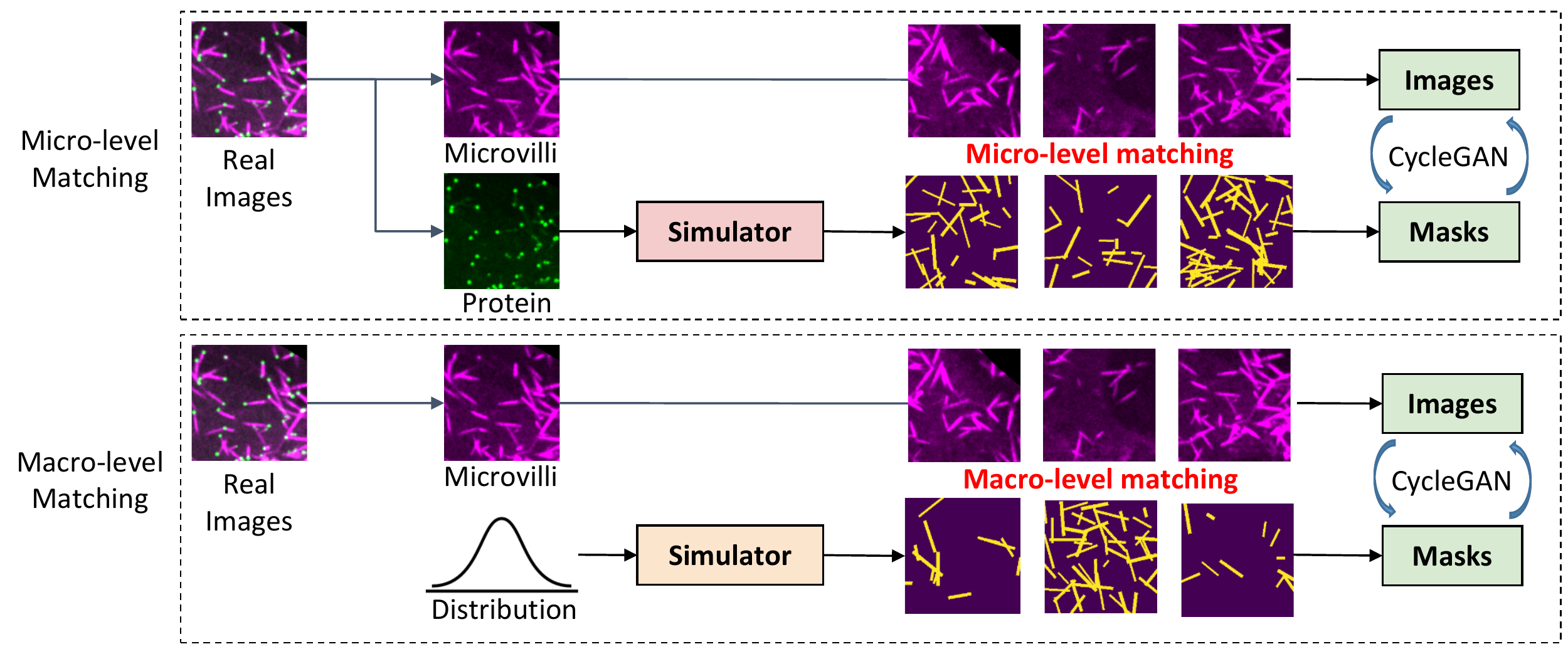}
\end{center}
   \caption{This figure shows the general framework of the image synthesis. CycleGAN is the used to perform synthesis between real images and simulated masks. In micro-level matching, the green protein channel in fluorescence images is used to achieve the cell counting automatically, which provides the rough numbers of microvilli in the real images. Then, the corresponding simulated masks with the same numbers of sticks are generated from the simulator when forming a mini-batch for training. In macro-level matching, the numbers of sticks in the simulated masks are randomly generated from the prior global distribution, without mini-batch level correspondence.}
\label{fig:ovewview}
 \end{figure}

\section{METHODS}

Our proposed unsupervised segmentation method consists of two parts: (1) image synthesis, and (2) segmentation. In image synthesis, our goal is to synthesize realistic looking images from the simulated masks. Then, the paired synthetic images and masks are used to train another segmentation network. \textbf{Note that, no manual annotations are used in our training either for CycleGAN or U-Net, as an unsupervised framework.}

\begin{figure}
\begin{center}
\includegraphics[width=1\linewidth]{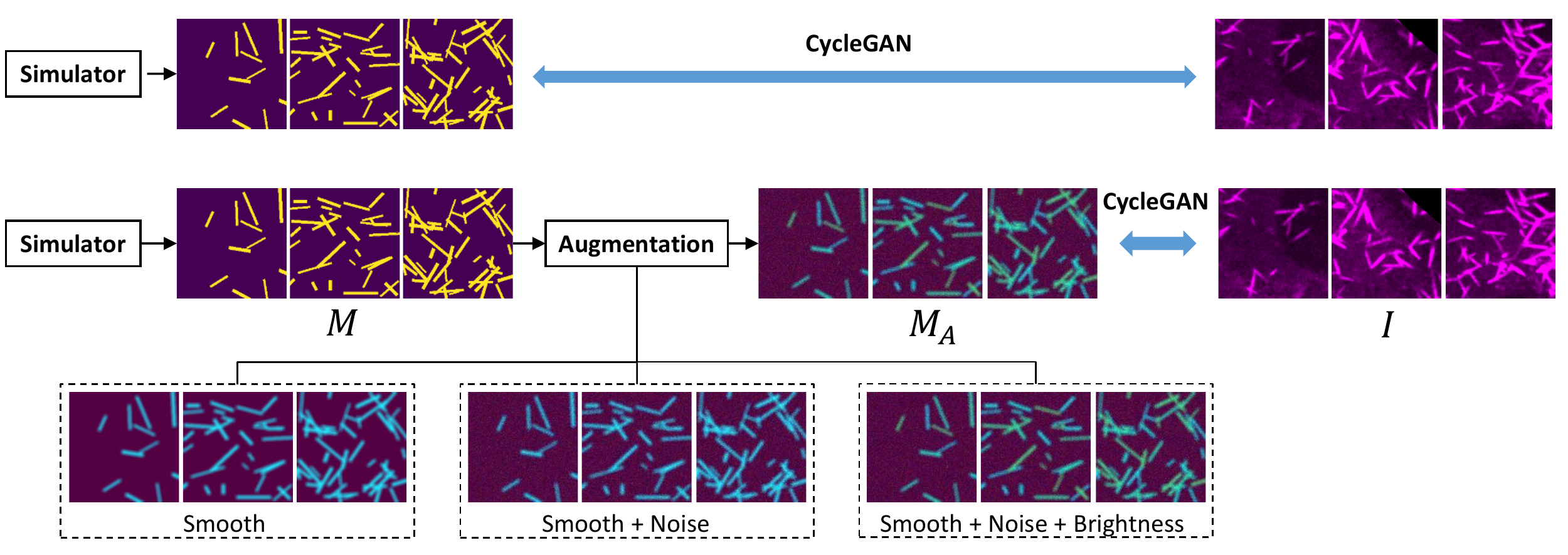}
\end{center}
   \caption{This figure demonstrates four experimental designs of providing different augmentations of masks for training the CycleGAN. The first design employs the binary mask directly, while the remaining three designs utilize different augmentation strategies.}
\label{fig:augment}
 \end{figure}
 
\subsection{Cycle-Consistent Image Synthesis}

The CycleGAN~\cite{zhu2017unpaired} is used to generate our synthetic training data. As the standard CycleGAN implementation, generators and discriminators are used to transfer the styles between two image modalities. The role of the generators is to convert the real images to another domain, which are typically called "fake" images. The discriminators then judge if a given image is real or fake. The CycleGAN model design creatively forms the entire learning process as a cycle-consistent loop, where the reconstructed fake images after two generators should be close to the original real images. In each branch, the generator tries to generate realistic images, while the discriminator try to distinguish the fake images from the real ones. 

In our unsupervised image segmentation framework (Fig.~\ref{fig:ovewview}), the CycleGAN is employed to synthesize segmentation masks (or called "annotations") from the real images $I$, and synthesize realistic looking images from simulated segmentation masks $M$. In the ideal case, the trained generator $G_{I->M}$ can be directly used as a segmentation network to segment new images. However, the quality of synthesis between real images and clean binary masks is typically not optimal since the underlying Poisson distribution of the binary masks is not a realistic distribution in real images~\cite{ihle2019udct}. Moreover, the optimization of the KL divergence for training discriminators is more difficult to converge~\cite{gadermayr2019generative} using clean binary masks. Therefore, the Gaussian smoothing, random noise, and brightness variations are used to generate augmented masks $M_A$ in additional to the simulated mask images $M$ for better synthetic performance (Fig.~\ref{fig:augment}). Then, the trained generator $G_{M_A->I}$ will provide us unlimited fake but realistic looking images $I_F$ from the simulated and augmented masks $M_A$. Eventually, the fake images $I_F$ and the clean binary masks $M_A$ (before augmentation) are used to train another independent segmentation network (see \textbf{Image Segmentation} section).

\subsection{Micro- and Macro-level Matching}
Compared with traditional pixel-to-pixel conditional GAN design which needs pixel-level correspondence between two modalities, the CycleGAN does not need paired images for training. However, the superior synthetic segmentation performance is typically achieved if the distributions of the number of objects in real image modality and annotation modality are roughly matched~\cite{ihle2019udct,gadermayr2019generative}, named as "macro-level" matching. However, no studies have explored the level of matching in the middle of pixel-level and macro-level. In this study, we proposed the idea called "micro-level" matching, which matches the number of objects in each mini-batch (Fig.~\ref{fig:ovewview}). For example, if a real image has roughly 21 microvilli, we will provide a simulated mask with the same 21 sticks, when forming the mini-batch. Then, the next question is that how can we get rough number of objects from the real images. In this study, we utilize the multi-channel nature of fluorescence microscopy to split the microvilli marker mCherry-Espin (magenta color objects) and microvilli tip marker EGFP-EPS8 (green color objects). Using the simple intensity thresholding based cell counting algorithm~\cite{refai2003automatic}, the rough number of protein objects are easily achieved. The numbers are then used as the rough number of microvilli to simulate the corresponding mask files with the same number of objects, as the micro-level matching. Note that we only match the number of the objects in the micro-level matching, where the spatial distribution of the objects is still random.

\begin{figure}
\begin{center}
\includegraphics[width=1\linewidth]{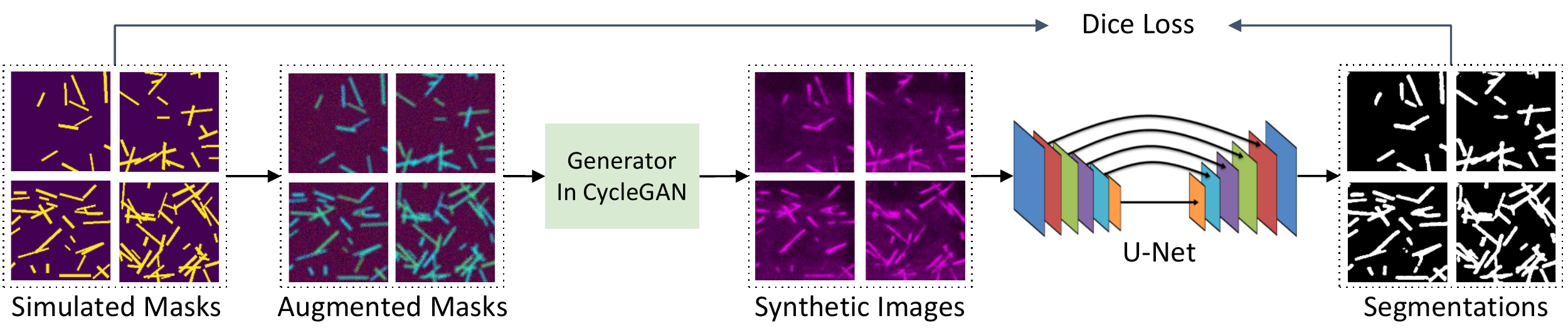}
\end{center}
   \caption{This figure shows the segmentation pipeline. The input images of the U-Net model are the synthetic fake images from trained CycleGAN's generator, while the annotations the simulated masks. Note that, no manual annotations are used in our training either for CycleGAN or U-Net, as an unsupervised framework. }
\label{fig:segment}
 \end{figure}

\subsection{Image Segmentation}
U-Net~\cite{ronneberger2015u} is employed as the segmentation backbone network, which is a fully convolutional neural network and widely used in image segmentation tasks. The segmentation part of our framework is shown in Fig.~\ref{fig:segment}. In our proposed unsupervised segmentation framework, the input images of U-Net are the fake microvilli images, which are generated from the simulated masks using $G_{M->I}$ or $G_{M_A->I}$ from trained CycleGAN. Then the Dice loss function is calculated by comparing the predicted segmentation with the simulated binary masks. Traditional deep neural network typically needs a large number of annotated images to train a segmentation network. Using our design, however, we can generate unlimited number of training data to train the segmentation network without any manual annotation efforts.

\section{Data and Experimental Design}

\subsection{Microvilli Images}
Twelve microvilli images acquired using fluorescence microscopy were used as training data, where each image had $\approx 900 \times 900$ pixels with pixel resolution 1.1 $\mu$m. Then, 500 image patches with $128 \times 128$ pixels were randomly sampled from the twelve images to train the CycleGAN as the real images. Then, another independent microvilli video with 20 frames was used as testing data to evaluate the performance of the proposed unsupervised segmentation methods. Each frame has $256 \times 256$ pixels with pixel resolution 1.1 $\mu$m. All microvilli in each frame were densely annotated manually by an experienced biologist as the gold standard segmentation. The video of microvilli frames and manual annotations are presented in the \textbf{supplementary materials:} \href{https://github.com/iamliuquan/GAN\_based\_segmentation}{https://github.com/iamliuquan/GAN\_based\_segmentation}.

\subsection{Experimental Design}
In order to test if micro-level matching can improve the unsupervised segmentation performance, we performed experiments using both macro-level and micro-level matching. For micro-level matching, the number of sticks of each images was obtained by automatically counting the number of green proteins. For macro-level matching, the number of sticks for each image was randomly sampled from a uniform distribution (range from 11 to 63), according to distribution of proteins. 

As shown in Fig~\ref{fig:augment}, we have four different augmentation settings to generate the simulated masks in annotation domain: \\
\noindent\textbf{Binary masks:} The binary masks were directly simulated as the images in the annotation domain, without any augmentation. Based on~\cite{meenderink2019actin} and the prior biological knowledge of microvilli, the width of each microvilli was simulated between 2 to 5 $\mu$m, while the length was simulated between 10 to 50 $\mu$m. As the pixel resolution of all our images was 1.1 $\mu$m, we randomly generated sticks with 2 to 4 pixels width and 9 to 45 pixels length from uniform distribution. \\
\noindent\textbf{Gaussian Smoothing:} The first augmentation was Gasussian smoothing, where a Gaussian filter with kernal size of $5 \times 5$ was applied to the binary masks. \\
\noindent\textbf{Random noise:} Upon the Gaussian smoothing, the random Gaussian noise was further applied to the entire mask image. The values of random noise ranged from 0 to 255 follows Gaussian distribution.\\
\noindent\textbf{Different brightness:}  To further introduce the global intensity variations, random intensity values (200 to 255) were assigned to each stick in binary masks, where maximum foreground intensity value was 255.

To improve the segmentation performance, CycleGAN was employed to synthesis cell images for U-Net model training. In our experiment, CycleGAN is used to learn the mapping from simulated masks to real microvilli cell images. We built up dataset in these two domains as CycleGAN model's input. Our CycleGAN model was trained for 60 epochs. According to the training loss, generator trained for 50 epochs shows the best performance. Generator trained in CycleGAN will be used to synthesis microvilli cell images based on simulated mask images. 

CycleGAN model cannot cover all details using original frames as input which has too many cells. For both CycleGAN and U-Net, the input images were all with $128 \times 128$ resolution cropped from original frames, and then resized to $256 \times 256$ during training. When applying trained U-Net on testing microvilli images, each testing image was first split to four $128 \times 128$ images, and the final segmentation were achieved by concatenating the corresponding four predictions back to the original resolution. The Dice results were calculated in the original $256 \times 256$ resolution for testing images. The CycleGAN and U-Net were deployed on a computer with GeForce GTX 1060 Graphic Card with 6 GB memory. To get better synthesised data and avoid over-fitting, the CycleGAN was trained with 50 epochs and the U-Net was trained with 10 epochs for all experiments. According to the prediction performance, U-Net has the best performance after 10 epochs. The results from the last epochs were reported in this paper.

\begin{figure}
\begin{center}
\includegraphics[width=1\linewidth]{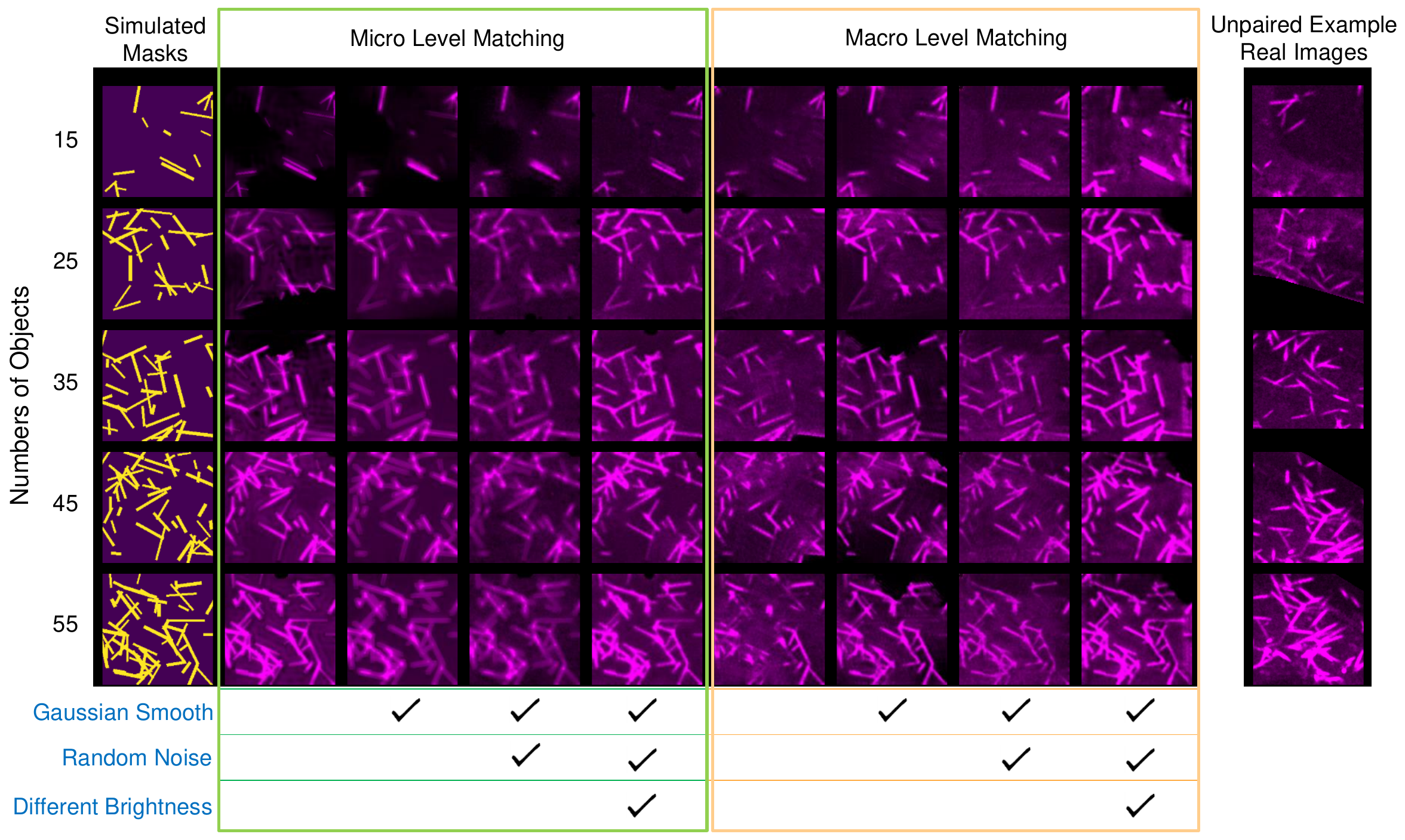}
\end{center}
   \caption{The synthesis results of different experimental designs are provided in this figure. The first column is the initial simulated masks with different numbers of objects (sticks). The middle columns exhibit the synthetic images from the masks using different augmentation strategies. The last column are five randomly selected real images, which are unpaired to the masks in CycleGAN framework.}
\label{fig:synthesis}
 \end{figure}

\begin{figure}
\begin{center}
\includegraphics[width=1\linewidth]{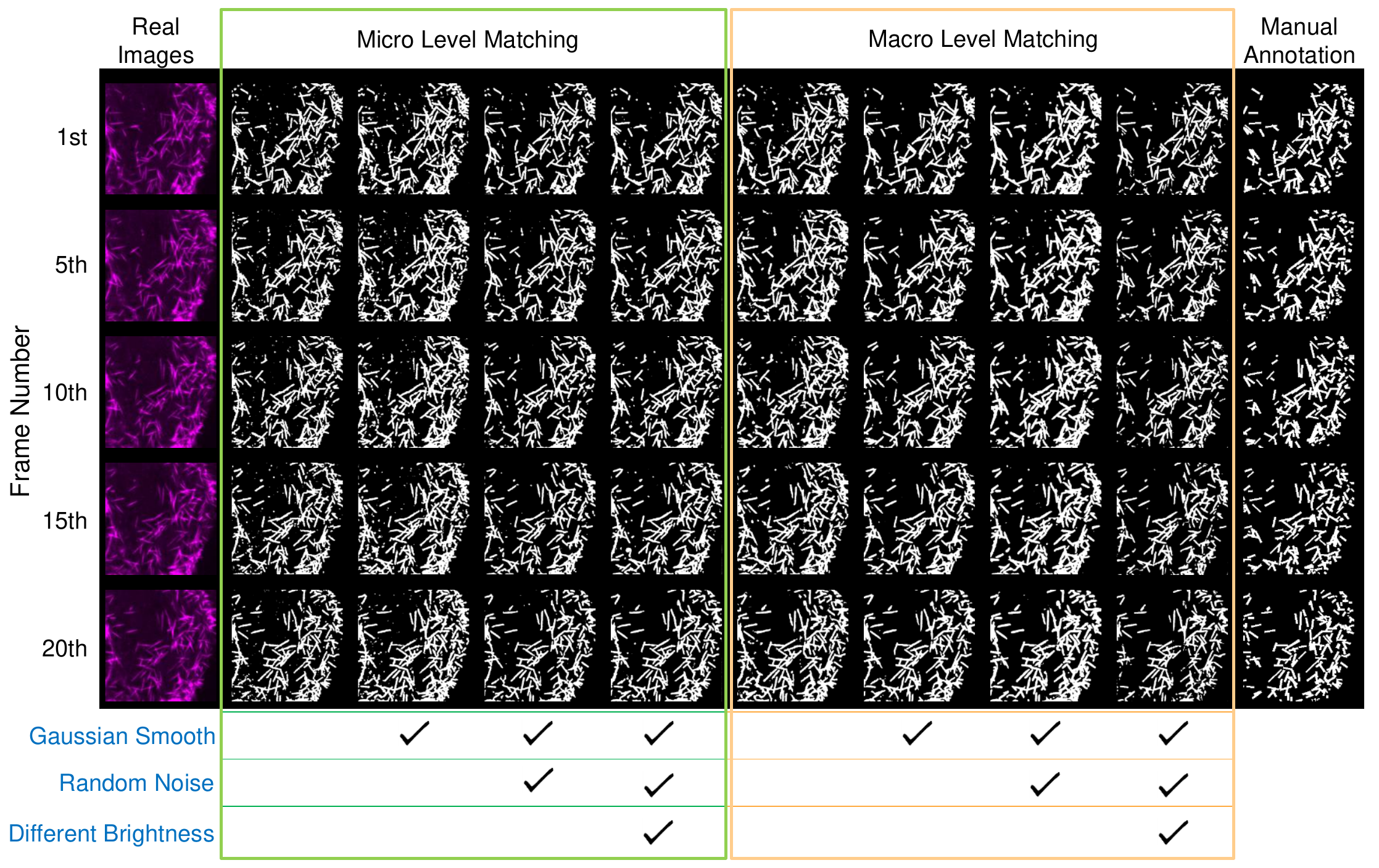}
\end{center}
   \caption{The final segmentation results from U-Net are presented in this figure. Each row shows the segmentation results at different frames in a microvilli video (the video of microvilli frames and our unsupervised segmentation results are presented in the \textbf{supplementary materials:} \href{https://github.com/iamliuquan/GAN\_based\_segmentation}{https://github.com/iamliuquan/GAN\_based\_segmentation}.}
\label{fig:mask}
\end{figure}

\setlength{\tabcolsep}{3pt}
\begin{table*}
\caption{The average Dice values of different experiments.}
\centering
\begin{tabular}{l@{}c@{\ \ }cccccccccccc}
\toprule
Exp.  &Smooth & Noise &Bright.  &$D_{(w=1)}$ & $D_{(w=2)}$ & $D_{(w=3)}$ & $D_{(w=4)}$ & $D_{(w=5)}$
  \\
\midrule
\multirow{4}{0.8in}{Micro-level matching} & & & &  0.3818 & 0.4860 & 0.5459 & 0.5605 & 0.5628 \\
& \checkmark  & & &  0.3650 & 0.4691 & 0.5301 & 0.5535 & 0.5667 \\
& \checkmark  & \checkmark & & 0.3738 & 0.4783 & 0.5367 & 0.5511 & 0.5547 \\
& \checkmark  & \checkmark & \checkmark& 0.3810 & 0.4865 & 0.5479  & 0.5650 & 0.5730 \\
\midrule
\multirow{4}{0.8in}{Macro-level matching} & & &  & 0.3639 & 0.4717 & 0.5364 & 0.5583 & 0.5675 \\
& \checkmark  & & & 0.3811 & 0.4918 & 0.5557 & 0.5776 & 0.5888 \\
& \checkmark  & \checkmark & & \textbf{0.3902} & \textbf{0.4981} & \textbf{0.5607} & \textbf{0.5965} & \textbf{0.6169} \\
& \checkmark  & \checkmark & \checkmark & 0.3894 & 0.4903 & 0.5467 & 0.5615 & 0.5631\\
\bottomrule
\end{tabular}
\\``$D$'' indicate the Dice score, $w$ means the width of the ground truth.
\label{table:res1}
\end{table*}
\setlength{\tabcolsep}{1.4pt}

\section{Results}

Considering both micro- and macro-level matching with different augmentation strategies, we performed eight experiments by training eight different CycleGAN networks. The qualitative results of image synthesis from eight different CycleGAN networks are provided in Fig.~\ref{fig:synthesis}. 

Then, the synthetic images were used to train eights different U-Net models using synthetic training image patches and applied to the real testing images. For testing images, the manual annotation was performed by tracking center line fragments of each microvillus (annotated by experienced biologist) since the traditional contour based annotations were extremely difficult on the tiny sub-cellular structures. To evaluate the segmentation results, we assigned different width to the manual segmentation and reported the results in Table.~\ref{table:res1}. The corresponding qualitative results of segmentation are provided in Fig.~\ref{fig:mask}. According to microvilli cell's biologic characteristic, manual annotation images are presented with width=3. From the results, the macro-level matching with Gaussian smoothing and random noise achieved the best performance across different widths of manual annotation. The micro-level matching did not improve the segmentation performance. Micro-level pairing can achieve higher accuracy on training dataset because its pairing is detailed to fit training dataset properties. Macro-level pairing is more robust. 
U-Net model trained by macro-level pairing performs better than micro-level pairing on new dataset. The standard Dice similarity coefficient metrics were used to evaluate different methods. The video of microvilli frames and our unsupervised segmentation results are presented in the \textbf{supplementary materials:} \href{https://github.com/iamliuquan/GAN\_based\_segmentation}{https://github.com/iamliuquan/GAN\_based\_segmentation}.

\section{Conclusion}

In this study, we proposed the first deep learning solution to enable unsupervised sub-cellular microvilli segmentation. Beyond the current standard macro-level matching strategy, we utilized the multi-channel nature of fluorescence microscopy to enable the micro-level matching of the number of objects in each mini-batch without introducing new human annotation efforts. From the experimental results, we conclude that the micro-level matching of object numbers at the mini-batch level did not lead to better segmentation performance. From the comprehensive analyses of introducing noise, smoothness and brightness, the Gaussian smoothing and random noise on the simulated annotations with macro-level matching resulted in the best microvilli segmentation performance.

\bibliography{report} 

\begin{thebibliography}{10}

\bibitem{wu1995live}
Wu, K., Gauthier, D., and Levine, M.~D., ``Live cell image segmentation,'' {\em
  IEEE Transactions on biomedical engineering}~{\bf 42}(1),  1--12 (1995).

\bibitem{pinidiyaarachchi2005seeded}
Pinidiyaarachchi, A. and W{\"a}hlby, C., ``Seeded watersheds for combined
  segmentation and tracking of cells,'' in [{\em International Conference on
  Image Analysis and Processing}{\nolinebreak\hspace{0.1em}]},   336--343,
  Springer (2005).

\bibitem{ragothaman2016unsupervised}
Ragothaman, S., Narasimhan, S., Basavaraj, M.~G., and Dewar, R., ``Unsupervised
  segmentation of cervical cell images using gaussian mixture model,'' in [{\em
  Proceedings of the IEEE conference on computer vision and pattern recognition
  workshops}{\nolinebreak\hspace{0.1em}]},   70--75 (2016).

\bibitem{lesko2010live}
Lesk{\'o}, M., Kato, Z., Nagy, A., Gombos, I., Torok, Z., Vigh~Jr, L., and
  Vigh, L., ``Live cell segmentation in fluorescence microscopy via graph
  cut,'' in [{\em 2010 20th International Conference on Pattern
  Recognition}{\nolinebreak\hspace{0.1em}]},   1485--1488, IEEE (2010).

\bibitem{moen2019deep}
Moen, E., Bannon, D., Kudo, T., Graf, W., Covert, M., and Van~Valen, D., ``Deep
  learning for cellular image analysis,'' {\em Nature methods} ,  1--14 (2019).

\bibitem{zhang2017deeppap}
Zhang, L., Lu, L., Nogues, I., Summers, R.~M., Liu, S., and Yao, J., ``Deeppap:
  deep convolutional networks for cervical cell classification,'' {\em IEEE
  journal of biomedical and health informatics}~{\bf 21}(6),  1633--1643
  (2017).

\bibitem{zhu2017unpaired}
Zhu, J.-Y., Park, T., Isola, P., and Efros, A.~A., ``Unpaired image-to-image
  translation using cycle-consistent adversarial networks,'' in [{\em
  Proceedings of the IEEE international conference on computer
  vision}{\nolinebreak\hspace{0.1em}]},   2223--2232 (2017).

\bibitem{goodfellow2014generative}
Goodfellow, I., Pouget-Abadie, J., Mirza, M., Xu, B., Warde-Farley, D., Ozair,
  S., Courville, A., and Bengio, Y., ``Generative adversarial nets,'' in [{\em
  Advances in neural information processing
  systems}{\nolinebreak\hspace{0.1em}]},   2672--2680 (2014).

\bibitem{huo2018adversarial}
Huo, Y., Xu, Z., Bao, S., Assad, A., Abramson, R.~G., and Landman, B.~A.,
  ``Adversarial synthesis learning enables segmentation without target modality
  ground truth,'' in [{\em 2018 IEEE 15th international symposium on biomedical
  imaging (ISBI 2018)}{\nolinebreak\hspace{0.1em}]},   1217--1220, IEEE (2018).

\bibitem{huo2018synseg}
Huo, Y., Xu, Z., Moon, H., Bao, S., Assad, A., Moyo, T.~K., Savona, M.~R.,
  Abramson, R.~G., and Landman, B.~A., ``Synseg-net: Synthetic segmentation
  without target modality ground truth,'' {\em IEEE transactions on medical
  imaging}~{\bf 38}(4),  1016--1025 (2018).

\bibitem{zhang2018translating}
Zhang, Z., Yang, L., and Zheng, Y., ``Translating and segmenting multimodal
  medical volumes with cycle-and shape-consistency generative adversarial
  network,'' in [{\em Proceedings of the IEEE conference on computer vision and
  pattern recognition}{\nolinebreak\hspace{0.1em}]},   9242--9251 (2018).

\bibitem{chen2019synergistic}
Chen, C., Dou, Q., Chen, H., Qin, J., and Heng, P.-A., ``Synergistic image and
  feature adaptation: Towards cross-modality domain adaptation for medical
  image segmentation,'' in [{\em Proceedings of the AAAI Conference on
  Artificial Intelligence}{\nolinebreak\hspace{0.1em}]},   {\bf 33},  865--872
  (2019).

\bibitem{ihle2019udct}
Ihle, S., Reichmuth, A.~M., Girardin, S., Han, H., Stauffer, F., Bonnin, A.,
  Stampanoni, M., V{\"o}r{\"o}s, J., and Forr{\'o}, C., ``Udct: Unsupervised
  data to content transformation with histogram-matching cycle-consistent
  generative adversarial networks,'' {\em bioRxiv} ,  563734 (2019).

\bibitem{gadermayr2019generative}
Gadermayr, M., Gupta, L., Appel, V., Boor, P., Klinkhammer, B.~M., and Merhof,
  D., ``Generative adversarial networks for facilitating stain-independent
  supervised and unsupervised segmentation: a study on kidney histology,'' {\em
  IEEE transactions on medical imaging}~{\bf 38}(10),  2293--2302 (2019).

\bibitem{dunn2019deepsynth}
Dunn, K.~W., Fu, C., Ho, D.~J., Lee, S., Han, S., Salama, P., and Delp, E.~J.,
  ``Deepsynth: Three-dimensional nuclear segmentation of biological images
  using neural networks trained with synthetic data,'' {\em Scientific
  reports}~{\bf 9}(1),  1--15 (2019).

\bibitem{julio2008image}
Julio, G., Merindano, M.~D., Canals, M., and Rall{\'o}, M., ``Image processing
  techniques to quantify microprojections on outer corneal epithelial cells,''
  {\em Journal of anatomy}~{\bf 212}(6),  879--886 (2008).

\bibitem{meenderink2019actin}
Meenderink, L.~M., Gaeta, I.~M., Postema, M.~M., Cencer, C.~S., Chinowsky,
  C.~R., Krystofiak, E.~S., Millis, B.~A., and Tyska, M.~J., ``Actin dynamics
  drive microvillar motility and clustering during brush border assembly,''
  {\em Developmental cell}~{\bf 50}(5),  545--556 (2019).

\bibitem{isola2017image}
Isola, P., Zhu, J.-Y., Zhou, T., and Efros, A.~A., ``Image-to-image translation
  with conditional adversarial networks,'' in [{\em Proceedings of the IEEE
  conference on computer vision and pattern
  recognition}{\nolinebreak\hspace{0.1em}]},   1125--1134 (2017).

\bibitem{refai2003automatic}
Refai, H., Li, L., Teague, T.~K., and Naukam, R., ``Automatic count of
  hepatocytes in microscopic images,'' in [{\em Proceedings 2003 International
  Conference on Image Processing (Cat. No.
  03CH37429)}{\nolinebreak\hspace{0.1em}]},   {\bf 2},  II--1101, IEEE (2003).

\bibitem{ronneberger2015u}
Ronneberger, O., Fischer, P., and Brox, T., ``U-net: Convolutional networks for
  biomedical image segmentation,'' in [{\em International Conference on Medical
  image computing and computer-assisted
  intervention}{\nolinebreak\hspace{0.1em}]},   234--241, Springer (2015).

\end{thebibliography}
\bibliographystyle{spiebib} 

\end{document}